\documentclass{elsarticle}

\usepackage[numbers]{natbib}

\usepackage{amssymb}
\usepackage{amsthm}
\usepackage{mathtools}
\usepackage{graphicx}
\usepackage{epstopdf}
\usepackage{caption}
\usepackage{amsmath}
\usepackage{algorithm}
\usepackage[noend]{algpseudocode}
\usepackage{multirow}
\usepackage{array}
\usepackage{calc}
\usepackage{tipa}
\usepackage{subfigure}
\usepackage{relsize}
\usepackage{setspace}
\usepackage[justification=centering]{caption}
\usepackage[table,xcdraw]{xcolor}

\usepackage{graphicx}

\usepackage{url}

\setcitestyle{square}

\journal{Special Issue on Meta-learning for Image/Video Segmentation}

\begin{document}

\begin{frontmatter}



\title{Cloud based Scalable Object Recognition from Video Streams using Orientation Fusion and Convolutional Neural Networks}


\author{Muhammad Usman Yaseen, Ashiq Anjum, Giancarlo Fortino, Antonio Liotta and Amir Hussain}

\address{Department of Computer Science, Comsats University, Pak}
\address{Department of Informatics, University of Leicester, Uk}
\address{Department of Informatics, University of Calabria, Italy}
\address{Faculty of Computer Science, Free University of Bozen-Bolzano, Italy}
\address{School of Computing, Edinburgh Napier University, UK}

\begin{abstract}
Object recognition from live video streams comes with numerous challenges such as the variation in illumination conditions and poses. Convolutional neural networks (CNNs) have been widely used  to perform intelligent visual object recognition. Yet, CNNs still suffer from severe accuracy degradation, particularly on illumination-variant datasets.
To address this problem, we propose a new CNN method based on orientation fusion for visual object recognition.  The proposed cloud-based video analytics system pioneers the use of bi-dimensional empirical mode decomposition to split a video frame into intrinsic mode functions (IMFs). We further propose these IMFs to endure Reisz transform to produce monogenic object components, which are in turn used for the training of CNNs.
Past works have demonstrated how the object orientation component may be used to pursue accuracy levels as high as 93\%. Herein we demonstrate how  a feature-fusion strategy of the orientation components leads to further improving visual recognition accuracy to 97\%. 
We also assess the scalability of our method, looking at both the number and the size of the video streams under scrutiny. We carry out extensive experimentation on the publicly available Yale dataset, including also a self generated video datasets, finding significant improvements (both in accuracy and scale), in comparison to AlexNet, LeNet and SE-ResNeXt, which are the three most commonly used deep learning models for visual object recognition and classification.

\end{abstract}




\end{frontmatter}


\section{Introduction}

Visual object recognition is an important component of a multimedia data analytics system, and aids in a number of applications including medical image processing, visual object tracking, interactive virtual reality games, among many others. To have a highly accurate visual object recognition system, multimedia applications have exploited knowledge from different domains including machine learning \cite{9084281}, computer vision \cite{nanni2017handcrafted}, distributed systems \cite{van2005grid} and pervasive computing \cite{KIANI20131107}.  

The most common challenges that the current visual object recognition systems experience include pose and illumination variations  \cite{hu2020face}, facial expressions, aging conditions \cite{li2018distance}, and crucially,  scalability \cite{steenberg2004clarens}. 


Convolutional Neural Networks (CNNs) have been used recently to perform visual object recognition on video datasets  \cite{babaee2018deep}. CNNs proved to be successful on a number of real-world applications including image reconstruction \cite{li2017hyperspectral}, scene classification \cite{nogueira2017towards}, and facial expression recognition \cite{lopes2017facial}. They have also been used for object detection and classification tasks on large video datasets  \cite{gu2018recent}. A great benefit of CNNs is that they  also have good generalization ability, and can be trained on large-scale video datasets, even on diverse classes. However, CNNs struggle to perform well on the more challenging datasets (as mentioned above), since their accuracy severely drops particularly with  expression- and illumination-variant datasets.

In order to achieve high visual object recognition accuracy on challenging datasets such as the Yale dataset, we propose an Empirical Mode Decomposition (EMD) implementation of CNNs. We split the input video dataset, consisting of images and videos, into its intrinsic mode functions (IMFs), by using EMD.  Reisz transform is then applied on the resulting IMFs to generate the monogenic components.  The local monogenic components including phase, orientation and amplitude are then analyzed to determine which of these components contribute the most to achieving a higher accuracy rate of the CNNs. 

\begin{figure*}[t]
  \centering
\includegraphics[width=\textwidth, height=5cm]{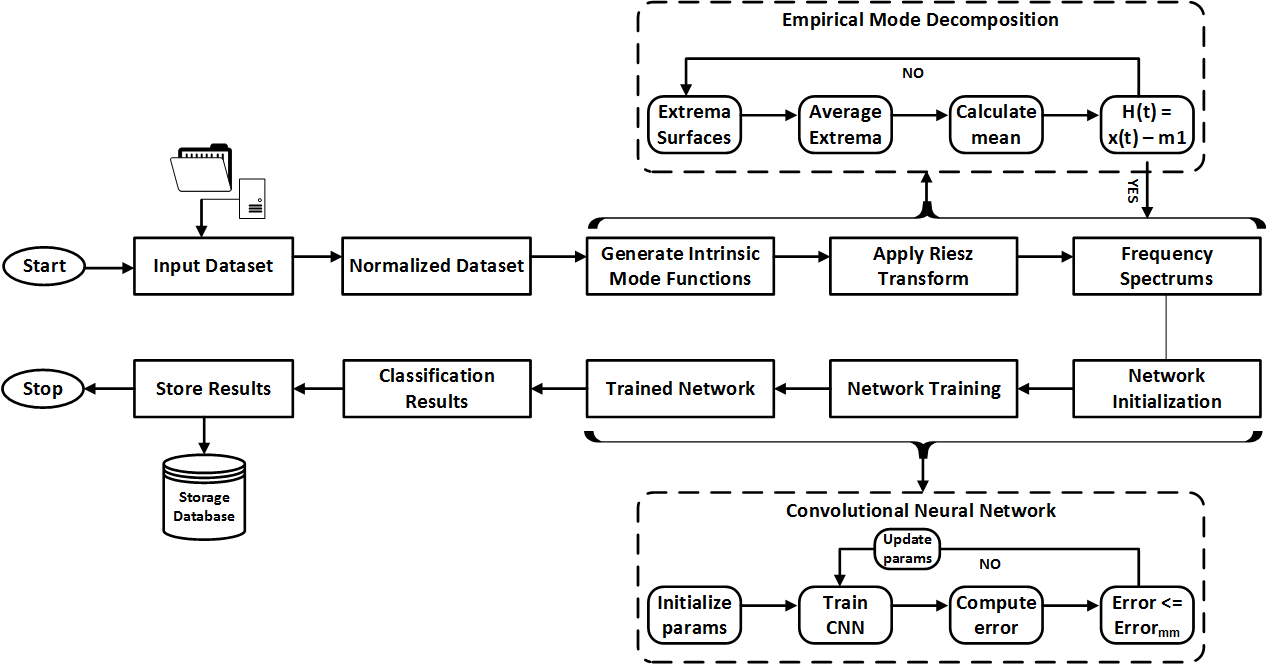}
\caption {Workflow of the proposed system: Upper half depicts the EMD procedure applied on the input dataset to generate IMFs, Lower half depicts the procedure of CNN training and classification}
\label{Architecture}
\end{figure*}

Figure \ref{Architecture} depicts the proposed visual object recognition system. The input dataset is first normalized to avoid the effects of unstable gradients during the training of the CNN. Empirical mode decomposition is then applied to generate intrinsic mode functions which pass through the Riesz transform to produce frequency spectra. The convolutional neural network is trained on these spectra and the trained model is used to perform classification. The classification results are stored in the database for further analysis.

As it will become evident from the experiments discussed in Sect. VI, 
the orientation component of the visual object is responsible for contributing  to the higher accuracy rates. Inspired by this fact, we further propose a feature-fusion strategy based on the orientation component of the IMFs. The top forty percent of the IMFs are relatively noise-free and contain higher frequency components. We have fused these higher IMFs into a single IMF \cite{zhao2020adaptive}, to produce a high-quality image, which leads to further improvements in visual object recognition rates. 

To achieve high scalability and video processing throughput \cite{hasham2011cms}, the proposed video analytics system has been deployed on a cloud infrastructure, based on the Spark distributed framework. The iterative map-reduce paradigm has been used to perform parallel training on multiple compute nodes. The parallel and distributed training model can process a large amount of data rapidly and efficiently. The underlying cloud infrastructure is iteratively tuned for maximum resource utilization, to support large-scale visual object recognition.

Our  novel contributions may be summarized as follows. 

\begin{itemize}
  \item  Firstly, this paper pioneers the use of empirical mode decomposition with CNNs, to improve visual object recognition accuracy on challenging video datasets. We study the orientation, phase and amplitude components and show their performance in terms of visual recognition accuracy. We show that the orientation component is a good candidate to achieve high object recognition accuracy, for illumination- and expression-variant video datasets.
  
  \item Secondly, we propose a feature-fusion strategy of the orientation components to further improve the accuracy rates. We  show that the orientation-fusion approach significantly improves the visual recognition accuracy, under challenging conditions. 
  
  \item Thirdly, we scale and optimize the underlying cloud-based infrastructure to improve the visual object recognition time of the system, so that it can be deployed on large video datasets.

\end{itemize}

The paper is organized as follows. Section II, reviews related work about visual object recognition and highlights their strengths and weaknesses. Section III, explains the proposed visual object recognition system. Section IV, details the architecture and implementation of proposed system. Evaluation results along with experimental setup are explained in Section V and VI, respectively. Section VII, draws conclusions  and pinpoints useful elements for future work.

\section{Related Work}
Researchers have been very active in applying shallow networks for object detection and classification problems. Object recognition systems based on shallow networks use hand-crafted features.


These are robust to noise and occlusion but require more computation time and resources. Yaseen et al. \cite{YASEEN2018286} proposed video analytics based on GPUs to speedup the feature extraction process. However, all the object recognition systems based on shallow networks produce high-dimensional feature vectors and are not suitable to work on large-scale data processing.

\begin{figure*}[t]
  \centering
\includegraphics[width=\textwidth, height=9cm]{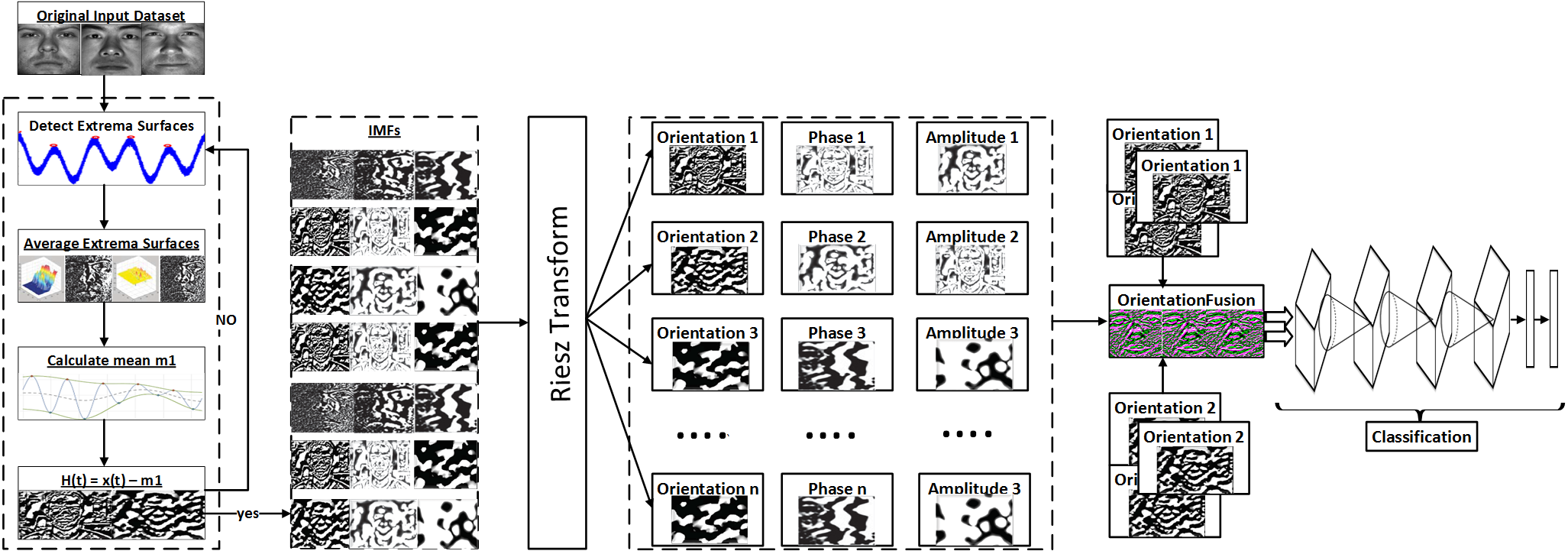}
\caption {The proposed approach: Riesz transform is applied on the IMFs to generate orientation spectra which are fused to perform classification}
\label{Approach}
\end{figure*}

Significant advances have been made possible with the advent of deep networks, which have increasingly been applied to  object detection and recognition, among a plethora of other usages. Shuiwang et al. \cite{6165309} proposed a 3D CNN for action recognition. They used both spatial and temporal dimensions to perform 3D convolutions. This helped to capture motion information present in adjacent video frames. 

Fully convolutional networks were used by Even et al. \cite{7478072} to perform semantic segmentation. Wright et al. \cite{4483511} proposed a face recognition system using sparse representation. They used multimodality data such as visual features and contextual information to recognize faces progressively, from publicly available datasets



Recent research showed that CNNs  can work well for images or video data, but mainly if these are of good quality. However, the accuracy of a CNNs severely degrades if these are applied on more challenging dataset, for instance  including illumination and noise challenges. Additionally, preprocessing techniques are also required to overcome these challenges, which results in additional processing cost.

Per contra, herein we have tackled this issue by using empirical mode decomposition \cite{EMD} and shifting the data from time domain to spatial-frequency domain. EMD has been used in the past to perform recognition and classification. 

Liu et al. \cite{liu2005directional} presented 2DEMD for edge detection. Yaseen et al. \cite{7877045} pioneered to utilize EMD on video data in a parallel and distributed system. 
However, all these works employed EMD in combination with shallow networks, whereas we explore the challenges and benefits of using EMD with  deep networks.

\begin{figure*}[!tbp]
  \centering
  \begin{minipage}[b]{0.47\textwidth}
    \includegraphics[width=\textwidth, height=3cm]{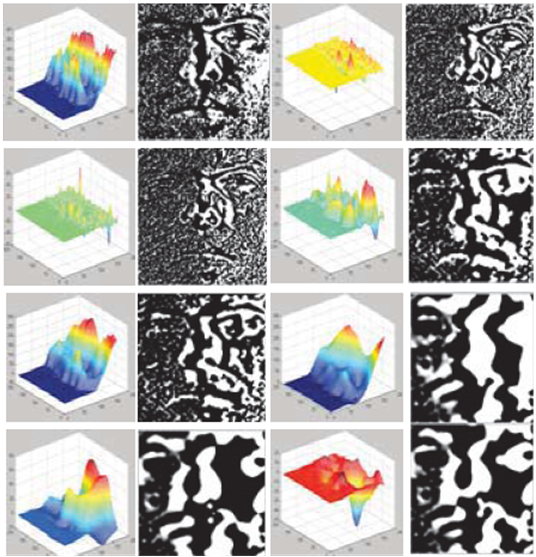}
    \caption{Averaged extrema surfaces used to generate intrinsic mode functions}
    \label{Averaged Extrema Surfaces}
  \end{minipage}
  \hfill
  \begin{minipage}[b]{0.49\textwidth}
    \includegraphics[width=\textwidth, height=3cm]{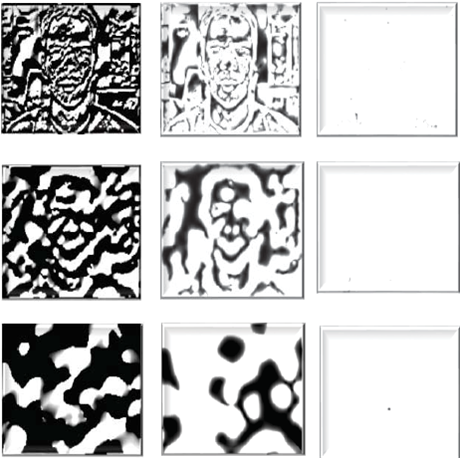}
    \caption{Amplitude, Phase and Orientation spectra of the first three IMFs}
    \label{Spectrums}
  \end{minipage}
\end{figure*}

\section{Object recognition approach and implementation}

This section describes our approach to visual object recognition,  as exemplified in Fig. \ref{Approach}. The input training dataset ``X" is represented by;

\begin{equation}
	``Training\ dataset\ X = {x_1,x_2, \dots ,x_n}" 
\end{equation}

Here, ``$x_1, x_2, \dots x_n"$ represent the individual subjects present in the training database. Each individual subject in the training database consists of a number of training samples,  given by:

\begin{equation}
\begin{split}
	``x_1 = {i_1,i_2, \dots ,i_n}\\ 
    x_2 = {i_1,i_2, \dots ,i_n}\\
    x_2 = {i_1,i_2, \dots ,i_n}\\
    \vdots \hspace{2cm} \vdots\\
    x_{n} = {i_1,i_2, \dots ,i_n}"
\end{split}
\end{equation}

Whereby, ``$ i_1, i_2, \dots ,i_n$" represent the individual images of each subject present in the training dataset. Each training sample $``i"$ from each individual subject $``x"$ undergoes through BEMD to have a decomposition into its frequency components. EMD generates these frequency components by a sifting process, in which the highest frequency components from the training sample are extracted in each cycle or mode. Each mode stores the high frequencies as an IMF. These IMFs are stored in the decreasing order of their frequencies, and the lowest IMF contains the lowest frequencies.

The sifting process first determines the extrema points from the training sample $``k(i, j)"$. The extrema points are then connected to form upper and lower envelops, respectively. An average of the upper and lower envelops is calculated to produce a mean envelop $``mean(i, j)"$, as shown in Fig.~\ref{Averaged Extrema Surfaces}, and is given by:

\begin{equation}
	``mean(i, j) = (eupper(i, j) + elower(i, j))/2"
\end{equation}

The mean envelop $``mean(i, j)"$ is then subtracted from the training sample $``k(i, j)"$ to produce $``T1"$, and is given by:

\begin{equation}
	``Tl_k = I(x, y) – m(x, y)" 
\end{equation}

The whole process is repeated till $``Tl_k" $ is a two-dimensional IMF. When the mean envelop $``mean(i, j)"$ reaches close to zero, this process is stopped; it otherwise  keeps on reiterating. The residual is obtained by removing the original training sample $``k(i, j)"$ from $``Tl_k" $. If the residual is represented by $``Res(i, j)"$, then

\begin{equation}
	``Res(i, j) = i(i, j) - Tl_k"
\end{equation}

In order to obtain the next IMF, the whole procedure is repeated on the residual $``Res(i, j)"$ by considering it as a training sample. The repetition of this process, on all the subsequent residuals results, in a number of IMFs, in the decreasing order of their frequencies is shown in Fig.~ \ref{Spectrums}. All the resultant IMFs and the residual can be grouped together to obtain the original training sample. Algorithm 1 depicts this whole procedure. 

\begin{algorithm}{}\label{alg1}
 \caption{Empirical Mode Decomposition}
  \begin{algorithmic}
\State \textbf{Input:} \\
 \hspace{0.5cm} Input Dataset  ${{x_1,x_2,…,x_n} }$, 192 $\times$ 168 image size \\
 \hspace{0.5cm} Width of each image W \\
 \hspace{0.5cm} Height of each image H \\
  \hspace{0.5cm} Number of iterations m \\
 \hspace{0.5cm} Number of IMFs n
\State \textbf{Output:}\\
      \hspace{0.5cm} result: IMFs \\
\While {!residue}
\State {Let the proto-IMF be \^x(x,y) = x(w,h)}
\While {IMF <= 3}
\While {!criteria}
\State {Identify local maxima and minima of (w,h)}
\State {Find envelop $e_{lower}(x,y)$ }
\State {Find envelop $e_{upper}(x,y)$ }
\State {Mean m(x,y)=($e_{upper}(x,y)$ = $e_{lower}(x,y)$ )/2}
\State {Extract detail $h_1$ = \^x(x,y) - m(x,y)}
\State {\^x(x,y) = $h_1$}
\State \textbf{end}
\EndWhile
\State {\^x(x,y) - $\sum_{j=1}^{3} h_j(x,y)$}
\State \textbf{end}
\EndWhile
\State \textbf{end}
\EndWhile
  \end{algorithmic}
\end{algorithm}

After obtaining all the required IMFs of the input data, the Riesz transform is applied to produce monogenic data. These  data aid in studying the local components of the input data.  The local components are calculated from each IMF.

\begin{algorithm}{}\label{alg:qt}
 \caption{Training weight vectors on local components}
  \begin{algorithmic}
\State \textbf{Input:} \\
 \hspace{0.5cm} Input Dataset  ${{x_1,x_2,…,x_n} }$, 192 $\times$ 168 image size \\
 \hspace{0.5cm} Output Target Label T 1-in-k vectors ${{y_1,y_2,…,y_t} }$ \\
 \hspace{0.5cm} Number of back-propagation epochs R \\
 \hspace{0.5cm} Number of convolution masks J \\
  \hspace{0.5cm} Activation function of convolution and subsampling g(.) \\
\State \textbf{Output:}\\
      \hspace{0.5cm} result: Recognition Labels \hspace{0.3cm} $f_{co}\Leftarrow FullyConnected$ \\
      \hspace{4.7cm} $result \Leftarrow Softmax(f_{co})$ 
\While {epoch r: 1 $\rightarrow$ R}
\While {Training image number x: 1 $\rightarrow$ X}
\State {Compute J hidden activation matrices $z_1, z_2,...,z_j$}
\begin{itemize} \State {g($x_{k,l} + w_{k,l} + B_{k,l}$)} \end{itemize}
\State {Downsample matrices $z_1, z_2,...,z_j$ by a factor of 2}
\begin{itemize} \State {g($\downarrow ^2$ $x_{k,l} + w_{k,l} + b_{k,l}$)} \end{itemize}
\State {Calculate weight and bias deltas}
\begin{itemize} \State {$\triangle W_{t,k} = lr \sum_{i=1}^{N} (x_i * D_i^h) + m\triangle W_{(t-1,k)}$} \end{itemize}
\begin{itemize} \State {$\triangle B_{t,k} = lr \sum_{i=1}^{N} D_i^h + m\triangle B_{(t-1,k)}$} \end{itemize}
\State {Calculate softmax activation vector 'a' }
\begin{itemize} \State {$l(i,x_{iT}) = M (e_i,f(x_{iT}))$} \end{itemize}
\State {Compute error $y_x -a$}
\State {Back propagate and update network weights}
\begin{itemize} \State { $W_{t+1} = W_t - \alpha\delta L(\theta_t)$} \end{itemize}
\State \textbf{end}
\EndWhile
\State \textbf{end}
\EndWhile
  \end{algorithmic}
\end{algorithm}
\begin{equation}
	``f_R(v) = I (v/v) \times f(v) = h_2(v) \times f(v)"
\end{equation}

\begin{figure}[t]
  \centering
\includegraphics[width=\textwidth/2, height=1.5cm]{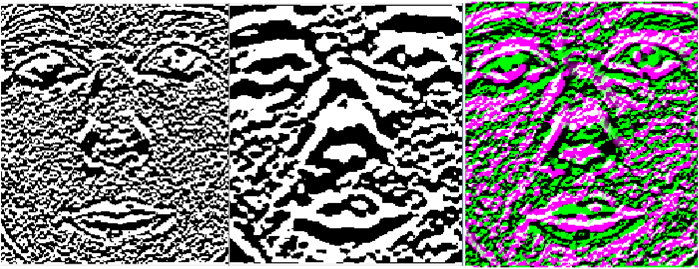}
\caption {First two orientation spectra and their fused orientation spectrum}
\label {OrientationFusion}
\end{figure}
Let ``$X_{amp}$" , ``$X_{pha}$" , ``$X_{ori}$"  be the amplitude, phase and orientation spectra of all the training samples present in the database. These can be represented as follows:

\begin{equation}
\begin{split}
	``X1_{amp} = {i_{amp1},i_{amp2},\dots,i_{amp\space n}}\\ 
    X1_{pha} = {i_{pha1},i_{pha2},\dots,i_{pha\space n}}\\
    X1_{ori} = {i_{ori1},i_{ori2},\dots,i_{ori\space n}}\\
    \vdots \hspace{2cm} \vdots \hspace{2cm} \vdots\\
    X34_{amp} = {i_{amp1},i_{amp2},\dots,i_{amp\space n}}\\ 
    X34_{pha} = {i_{pha1},i_{pha2},\dots,i_{pha\space n}}\\
    X34_{ori} = {i_{ori1},i_{ori2},\dots,i_{ori\space n}}"\\
\end{split}
\end{equation}
Here, ``$x_{amp}$" , ``$x_{pha}$ \dots $x_{ori}$"  represent the amplitude, phase and orientation spectra of individual subjects present in the training database. We created a fused orientation spectrum from the orientation spectra of higher IMFs. Since most of the illumination effects and noise present in the images resides in the lowest frequency bands; so we have discarded the lower IMFs, retaining only the higher IMFs, containing the high-frequency components. The fused spectrum is a composite spectrum which contains the elements from both the spectra. The fused spectrum merges the orientation spectra of higher IMFs into a single surface, adding in the meaning of the original orientation spectra. 

The fused spectrum shows the original orientation spectra overlaid in different color bands, as shown in Fig.~\ref{OrientationFusion}. The gray regions in the fused spectrum show where the original orientation spectra posses the same intensities. On the other hand, the colored regions show where the spectra have different intensities. These colored regions play an important role in enhancing the discriminative capabilities during the feature extraction process. It can be visualized from the figure that the fused intrinsic mode function contains significant information in terms of data points, which is what leads to further improvements in the accuracy rates. Let ``$XFused_{ori}$"  represents the fused orientation spectrum and ``*"  represents the fusion operation. The fused orientation spectrum of all the subjects is then given as:
\begin{equation}\label{my_sec_eqn}
``\triangle XFused_{ori} = \sum_{i=1}^{N} (x(i)_{ori1} * x(i)_{ori2})"
\end{equation}
The network is trained on these datasets separately in different experiments, and their effects are studied on the overall performance of the system, which helps explaining which dataset gives the best performance in terms of accuracy to discriminate and classify among different classes.

\begin{figure*}[t]
  \centering
\includegraphics[width=\textwidth, height=3.5cm]{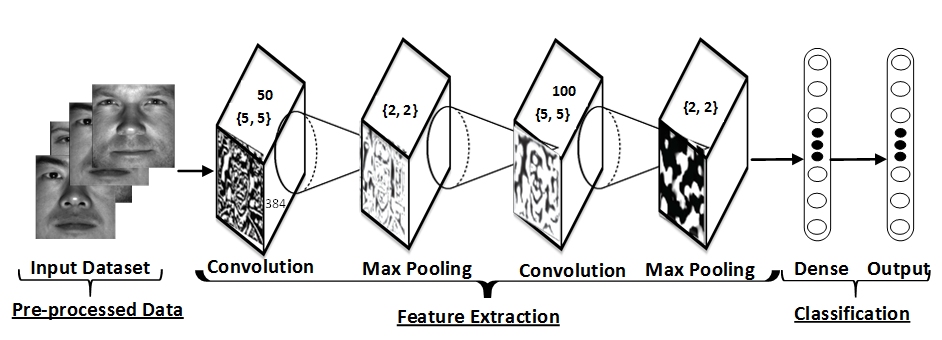}
\caption {Convolutional neural network architecture: Each convolution layer follows max pooling layer}
\label{CNN Model}
\end{figure*}

The convolutional layers and sub-sampling layers of the CNN used in our system are represented as:

\begin{equation}\label{my_sec_eqn}
``Convol_{i,j} = g(x_i,j * W_i,j + B_i,j)"
\end{equation}

\begin{equation}\label{my_sec_eqn}
``Subsamp_{i,j} = g(\downarrow x_i,j * w_i,j + b_i,j)"
\end{equation}

The weight and bias for sub-sampling layers are calculated as:

\begin{equation}\label{my_sec_eqn}
``\triangle W_{t,i} = lr \sum_{i=1}^{N} (x_i * D_i^h) + m\triangle W_{(t-1,i)}"
\end{equation}

\begin{equation}\label{my_sec_eqn}
``\triangle B_{t,i} = lr \sum_{i=1}^{N} D_i^h + m\triangle B_{(t-1,i)}"
\end{equation}

We have used ReLu as the activation function in our framework, which is represented by g(.) in the above equation. The weight and bias vectors are represented by $``W"$ and $``B"$ in the equations. The inputs are convolved with the weight vectors of the network with the help of a two-dimensional convolution operation represented by $``*"$ in the equations. The sub-sampling layer down-samples the given input. The range of ReLu activation functions goes over from 0 to infinity, and can model positive real numbers. It works much better for CNNs, as compared to the sigmoid function, because when the value of $``x"$ increases, the ReLu function does not vanish. 

The stochastic gradient descent and the momentum term used in the training of the network are given by:
\begin{equation}\label{my_sec_eqn}
``W_{t+1} = W_t - \alpha\delta L(\theta_t)"
\end{equation}

\begin{equation}\label{my_sec_eqn}
``V_{t+1} = \rho v_{t} - \alpha\delta L(\theta_t)"
\end{equation}
\begin{equation}\label{my_sec_eqn}
``W_{t+1} = W_t + V_{t+1}"
\end{equation}

The softmax layer, which is the last layer of the network, is given by:
\begin{equation}\label{my_sec_eqn}
``l(i,x_i T)=M(e_i , f(x_i T))" 
\end{equation}

The proposed visual object recognition system is compute intensive, as it is built upon CNNs that require large training times. We have optimized the code, and tuned the hyper-parameters as to perform training in a reasonable amount of time \cite{yaseendeep}. 

The initial parameters are initialized and the network configurations are loaded to start the training process as shown in Algorithm 2. The dataset is divided into a number of mini-batches, as loading the data into the memory at once would not be feasible. The size of the mini-batch is dependent on the settings of the network configuration. The mini-batches facilitate in tackling the memory requirement issue. A mini-batch of value 12 is used in the proposed system, which is selected on the basis of experimentation.


We have adopted the Local Response Normalization to aid in generalization. We have used max pooling in the pooling layer to perform sample based discretization. Max pooling decreases the dimensionality, reduces the number of parameters to learn, and also cuts down the overall computational cost. 

The input training samples are first filtered by 50 kernels having  dimensions and stride of 192 x 168 x 1 and 1 x 1, respectively. The subsequent and preceding layers have associated kernels to each other with a nonZeroBias. The max-pooling layer which is next to convolutional layers has a dimension of 2 x 2. All these layers end up onto a fully connected layer. The proposed CNN model architecture is shown in Fig.~\ref{CNN Model}.

\begin{figure}[t]
  \centering
\includegraphics[width= 10cm, height=4cm]{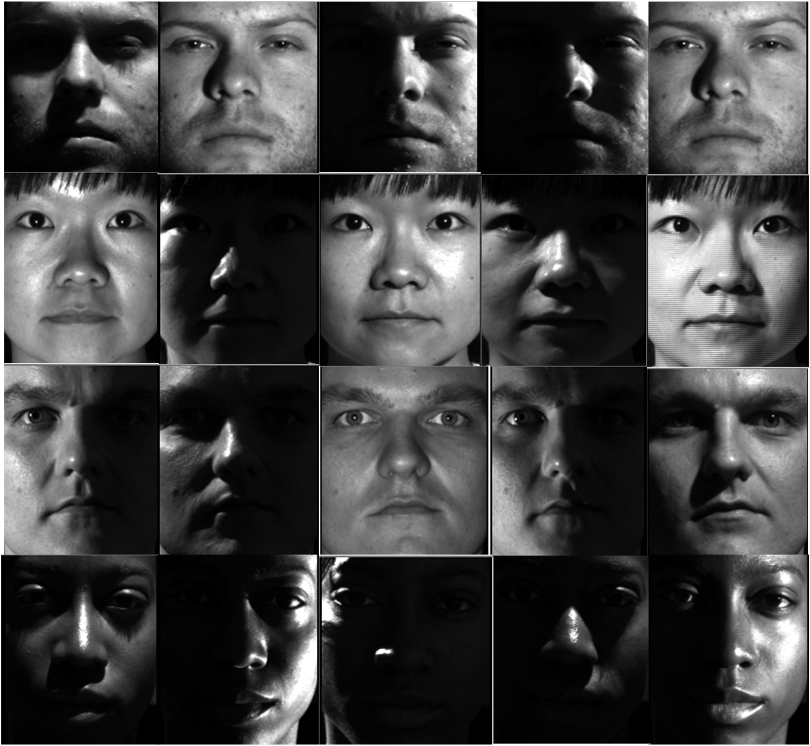}
\caption {Example faces from cropped Yale face dataset}
\label{Cropped Yale}
\end{figure}

\section{Experimental Setup}
The following performance characterization has been used for evaluation purposes: Accuracy (Acc), Precision (Pre), Recall (Rec) and F1 Score. The generated results are further discussed through the confusion matrix, in terms of FalseNegatives (FN), FalsePositives (FP), TruePositives (TP) and TrueNegatives (TN). Accuracy is defined as the percentage of correctly classified instances, i.e. (TP + TN)/(TP + TN + FP + FN).  



To measure the efficiency of the proposed system we have used the cropped Yale face dataset, in addition to our own self-generated dataset. The training and testing datasets are separated manually for validation purposes. The testing dataset contains samples from each subject from the training dataset with variations in expressions, pose and illumination. 

The images in the Yale database are captured at a resolution of 168x192 pixels, as shown in Fig.~\ref{Cropped Yale}. Every subject present in the database demonstrates illumination variations. It also demonstrates variations in expressions. The self-generated video dataset  contains the illumination, pose and facial expression challenges too, and is very similar to the Yale face database.

\begin{figure}[t]
  \centering
\includegraphics[width=\textwidth, height=8cm]{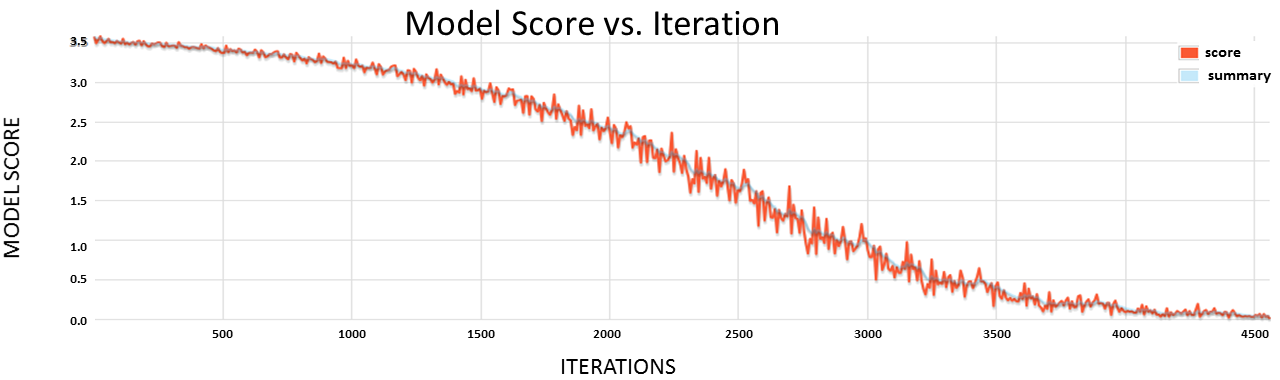}
\caption {Model Score vs. Iteration during CNN training}
\label {Model Score vs. Iteration}
\end{figure}

\begin{figure}[t]
  \centering
\includegraphics[width=\textwidth, height=7cm]{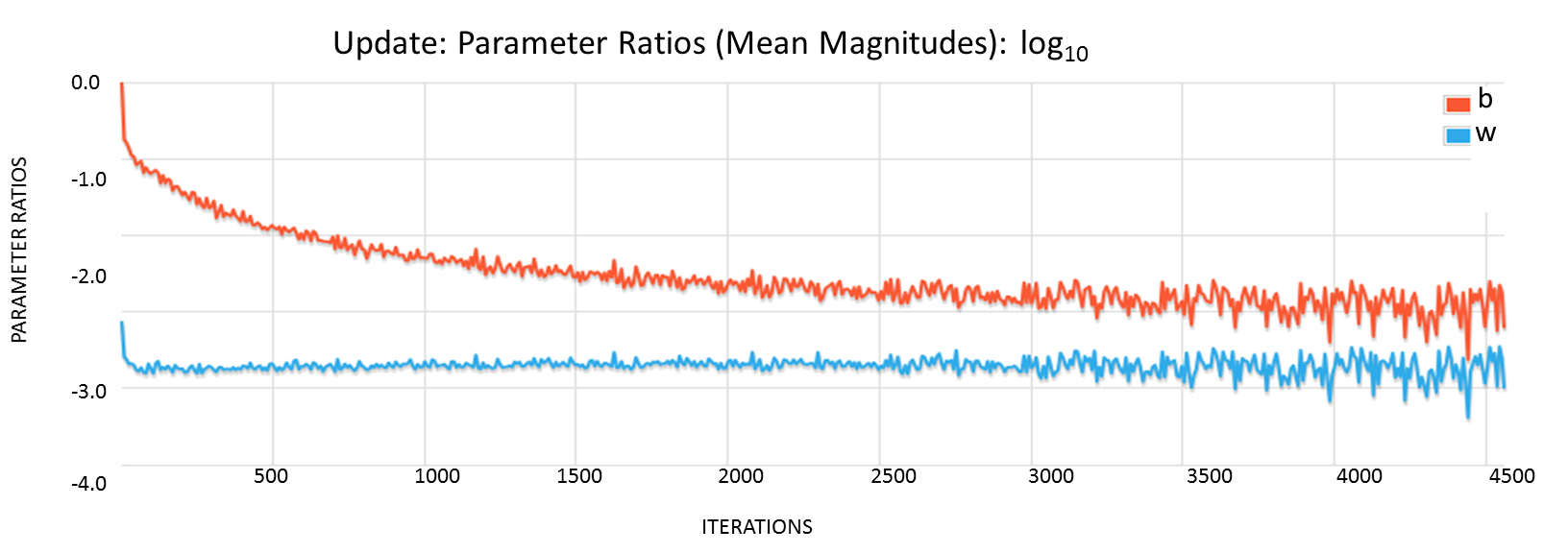}
\caption {Parameter Ratios during CNN training}
\label {Parameter Ratios}
\end{figure}

For evaluation purposes, we have compared the proposed system based on the two well-known models AlexNet and LeNet. The AlexNet model consists of 13 layers in total. There are five convolutional layers and three subsampling layers. Two of these subsampling layers precede LRN layers. On the other hand, the LeNet model has five layers. 



\section{Experimental Results}

The results and discussion of the proposed system are presented in this section, including three main topics. (i) Firstly, we describe the training of the proposed model and visualize the performance of the training parameters, during the model training. The visualization of weight vectors and other parameters during the training helps for proper parameter tuning. (ii) Secondly, we make a comparison of the proposed system with the two existing models and measure the improvements in terms of Accuracy, Recall and Precision. A discussion on the performance characterization of the resulting confusion matrix is also provided. (iii) Thirdly, we present the results about the system's scalability and the overall performance of our cloud based implementation.\\

\textbf {\textit {Deep Learning Model Training.}} Fig.~\ref{Model Score vs. Iteration} shows loss at increasing number of iterations per unit time. This is shown on the current mini-batch size. The graph shows a descending trend over multiple iterations over time. Especially a rapid decline in the graph is observed after the completion of 1500 iterations. Then, a sustained  decline continues up until  3500 iterations. The graph then keeps on decreasing until the value of the loss function approaches to zero. This converging trend in the loss function graph shows that the model parameters including network weights, learning rate and regularization are tuned properly.

We have selected the learning rate for the proposed system on the basis of a number of experiments. We have tested and visualized the model training on different learning rate values. The learning rate value of 0.0001 helped the network to converge more rapidly, as compared to 1e-4 and 1e-6. The normalization of data is performed properly with L2 normalization and stochastic gradient descent $ W_{t+1} = W_t - \alpha\delta L(\theta_t) $, which is depicted by the decreasing trend of the graph.   

Figure \ref{Parameter Ratios} depicts the parameter ratios, for each weight vector of each layer. These ratios are the mean magnitudes of the layer parameters. The mean magnitudes represent the mean or average value of the parameters on a number of iterations, which are shown on the y-axis of the graph. Figure \ref{Parameter Ratios} shows that the values remain between the suggested range i.e. -3.0 and -4.0 on a log10 chart, depicting an appropriate initialization of all network hyper-parameters. When the graph diverges away from the suggested range during the training, this indicates that the parameter initialization and selection is unstable and the model remained unable to learn the required distinguishing features from the training dataset. 

 \begin{figure}[t]
   \centering
\includegraphics[width=\textwidth, height=4cm]{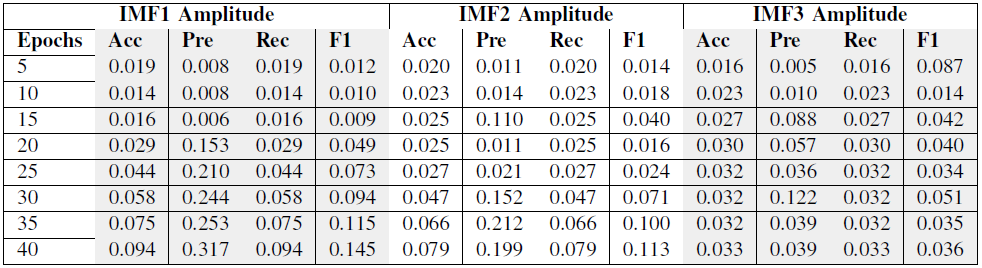}
 \caption {Performance Measures of the Amplitude Component}
 \label {amplitude}
 \end{figure}

 \begin{figure}[t]
   \centering
\includegraphics[width=\textwidth, height=4cm]{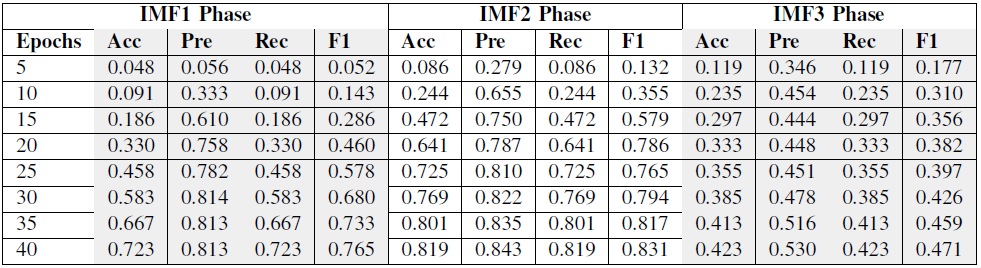}
 \caption {Performance Measures of the Phase Component}
 \label {phase}
 \end{figure}

We have also generated the histogram of layer parameters and layer updates. The layer updates are obtained when the learning rate, regularization and momentum ``$V_{t+1} = \rho v_{t} - \alpha\delta L(\theta_t)$" ($\rho$ is varied from 0.6 to 0.9 with 0.9 being the best for training) are applied. We have observed ``Gaussian Distribution" in both histograms, depicting that the network is free from the exploding gradient problem. We believe that this is due to the addition of gradient normalization in the network.\\

\textbf {\textit {Performance of the Deep Learning Model.}} The classifier's performance is measured using the following performance characterizations: Accuracy, Recall, Precision and F1 score. We have calculated these performance measures for the first three IMFs for amplitude, phase and orientation components, and made a comparison of these to evaluate the best performing components. The component which contributes the most in improving the accuracy of the classifier is then further used for fusion. The number of epochs during the training of the classifier have been varied from 5 to 40. We have also calculated the training time of classifier for each epoch to have an estimate of the total training time. 
Important results may also be derived by observing the confusion matrix (see below), where we have compared our approach to popular CNN models . 

Figure \ref{amplitude} shows the performance of the classifier for the amplitude component. The accuracy, recall, precision and F1 scores are tabulated for the first three IMFs. The number of epochs has been varied from 5 to 40. It can be seen from the table that the amplitude component could not perform better, and remained unable to classify the test patterns. Even as the number of epochs increased, it still provided poor performance. We believe that the reason for such a poor performance is due to the scarce availability of the data points in the amplitude component. As can be observed from Fig.~\ref{Spectrums},  the amplitude component does not provide a significant number of data points, which would be required to perform an accurate classification. With the reduction in the noisy components, it also discards the useful data points, which could have aided in performing a more accurate classification. This results in the drop of the overall accuracy rate of the classifier.

Table \ref{phase} shows the performance of the classifier for the phase component. The accuracy, recall, precision and F1 scores are, again, calculated for first three IMFs, with the number of epochs varying from 5 to 40. We can see that the phase component performed much better to classify the test patterns, as compared to the amplitude component. Increasing the number of epochs improved the accuracy of the classifier. Especially after 20 epochs the accuracy improved significantly, and at epoch number 40 it reached  an accuracy of 72\%. The improved performance rates were observed for precision, recall and F1 scores as well. One reason for the better performance could be the availability of much higher number of data points for the phase component, as compared to the amplitude component, as it can be seen from Fig.~\ref{Spectrums}. Although the phase component for the third intrinsic mode function does not contain enough data points, the first two were enough to have a descent classification rate.

 \begin{figure}[t]
   \centering
\includegraphics[width=\textwidth, height=4cm]{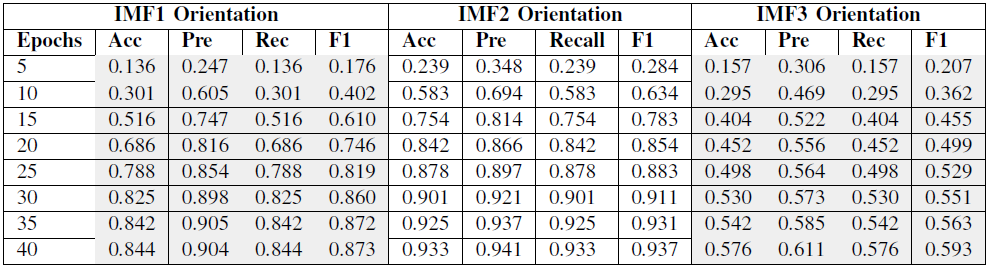}
 \caption {Performance Measures of the Orientation Component}
 \label {orientation}
 \end{figure}

The performance of the classifier for the orientation component is depicted in Figure \ref{orientation}. The same performance measures are tabulated for the first three IMFs of the orientation component. The number of epochs are, again, varied from 5 to 40 for these set of experiments. A significant amount of improvement in the overall accuracy rate of the classifier has been observed, as compared to both amplitude and phase components. Even at epoch number 10 the classification accuracy started at a reasonable rate of 0.3015 and kept on improving to 0.84 till the 40th epoch. The improved performance rates were observed for precision, recall and F1 scores as well. The precision was recorded to be 0.9047 at the 40th epoch. Similarly, the recall and F1 scores are recorded to be 0.84 and 0.93, respectively. These are much better rates that with the other two components. We figured out that these improvements are due to the maximum availability of data points, and minimum amount of presence of noisy frequencies in the orientation component. This can be seen clearly in Fig.~\ref{Spectrums}, whereby  even the third intrinsic mode function kept reasonable number of data points, which contributed towards the improvements in higher accuracy rates.

Inspired by these encouraging results, we have performed a feature-fusion strategy for the orientation components to further improve the accuracy rates. The feature-fusion strategy is performed on the higher intrinsic mode functions of the orientation component. The intrinsic mode functions are fused together in order to have a composite intrinsic mode function, which could hold the properties of both of the IMFs. The fused IMF is a numeric matrix which represents the combined orientation component and is used for classification. 

Table \ref{Orientation Fusion} shows the performance of the classifier on the fused intrinsic mode function. The number of epochs are varied from 5 to 40, and accuracy, precision, recall and F1 scores are recorded and tabulated in the table. It can be seen that the overall accuracy rate of the fused IMF is greater than all the previous accuracy figures of the amplitude, phase and orientation components. From epoch number 10 the classification accuracy was recorded to be 0.53, which is significantly better than the amplitude and phase and orientation. The accuracy kept on improving to 0.97, until the 40th epoch. Improvements were also observed for precision, recall and F1 scores. The precision was recorded to be 0.9806 at the 40th epoch. Similarly, the recall and F1 scores were recorded to be 0.97 and 0.98, respectively. We believe that these improvements are due to the further addition of data points in the orientation component, and a reduced presence of noisy frequencies. This can be visualized in Fig.~\ref{OrientationFusion}, where  the fused intrinsic mode function contains significant information in terms of data points, which leads to further improvements in higher accuracy rates.

\begin{table}[]
\centering
\caption{Performance Measures of Orientation Fusion}
\label{Orientation Fusion}
\begin{tabular}{|l|l|l|l|l|}
\hline
\multicolumn{5}{|c|}{\textbf{Orientation Fusion}}                                                                                                                                                                                                                                          \\ \hline
\textbf{Epochs} & \cellcolor[HTML]{EFEFEF}{\color[HTML]{000000} \textbf{Accuracy}} & \cellcolor[HTML]{EFEFEF}{\color[HTML]{000000} \textbf{Precision}} & \cellcolor[HTML]{EFEFEF}{\color[HTML]{000000} \textbf{Recall}} & \cellcolor[HTML]{EFEFEF}{\color[HTML]{000000} \textbf{F1 Score}} \\ \hline
5               & \cellcolor[HTML]{EFEFEF}{\color[HTML]{000000} 0.189}            & \cellcolor[HTML]{EFEFEF}{\color[HTML]{000000} 0.531}             & \cellcolor[HTML]{EFEFEF}{\color[HTML]{000000} 0.189}          & \cellcolor[HTML]{EFEFEF}{\color[HTML]{000000} 0.279}            \\ \hline
10              & \cellcolor[HTML]{EFEFEF}{\color[HTML]{000000} 0.533}            & \cellcolor[HTML]{EFEFEF}{\color[HTML]{000000} 0.767}             & \cellcolor[HTML]{EFEFEF}{\color[HTML]{000000} 0.533}          & \cellcolor[HTML]{EFEFEF}{\color[HTML]{000000} 0.629}            \\ \hline
15              & \cellcolor[HTML]{EFEFEF}{\color[HTML]{000000} 0.804}            & \cellcolor[HTML]{EFEFEF}{\color[HTML]{000000} 0.861}             & \cellcolor[HTML]{EFEFEF}{\color[HTML]{000000} 0.804}          & \cellcolor[HTML]{EFEFEF}{\color[HTML]{000000} 0.831}            \\ \hline
20              & \cellcolor[HTML]{EFEFEF}{\color[HTML]{000000} 0.910}            & \cellcolor[HTML]{EFEFEF}{\color[HTML]{000000} 0.927}             & \cellcolor[HTML]{EFEFEF}{\color[HTML]{000000} 0.910}          & \cellcolor[HTML]{EFEFEF}{\color[HTML]{000000} 0.918}            \\ \hline
25              & \cellcolor[HTML]{EFEFEF}{\color[HTML]{000000} 0.954}            & \cellcolor[HTML]{EFEFEF}{\color[HTML]{000000} 0.960}             & \cellcolor[HTML]{EFEFEF}{\color[HTML]{000000} 0.954}          & \cellcolor[HTML]{EFEFEF}{\color[HTML]{000000} 0.957}            \\ \hline
30              & \cellcolor[HTML]{EFEFEF}{\color[HTML]{000000} 0.967}            & \cellcolor[HTML]{EFEFEF}{\color[HTML]{000000} 0.970}             & \cellcolor[HTML]{EFEFEF}{\color[HTML]{000000} 0.967}          & \cellcolor[HTML]{EFEFEF}{\color[HTML]{000000} 0.969}            \\ \hline
35              & \cellcolor[HTML]{EFEFEF}{\color[HTML]{000000} 0.977}            & \cellcolor[HTML]{EFEFEF}{\color[HTML]{000000} 0.979}             & \cellcolor[HTML]{EFEFEF}{\color[HTML]{000000} 0.977}          & \cellcolor[HTML]{EFEFEF}{\color[HTML]{000000} 0.978}            \\ \hline
40              & \cellcolor[HTML]{EFEFEF}{\color[HTML]{000000} 0.979}            & \cellcolor[HTML]{EFEFEF}{\color[HTML]{000000} 0.980}             & \cellcolor[HTML]{EFEFEF}{\color[HTML]{000000} 0.979}          & \cellcolor[HTML]{EFEFEF}{\color[HTML]{000000} 0.980}              \\ \hline
\end{tabular}
\end{table}

The overall accuracy of the system with the fused features is recorded to be 0.9794. The precision is recorded to be 0.9806, which shows that the proposed system is accurate as well as precise. The recall is 0.9794, and the F1 score is observed as 0.98. Most of the test samples from all the subjects were classified correctly by the classifier. There were few samples from some subjects that were miss-classified. We believe that this  was due to the severe illumination effect in the test samples. 

\begin{figure*}[!tbp]
  \centering
  \begin{minipage}[b]{0.47\textwidth}
    \includegraphics[width=\textwidth, height=6cm]{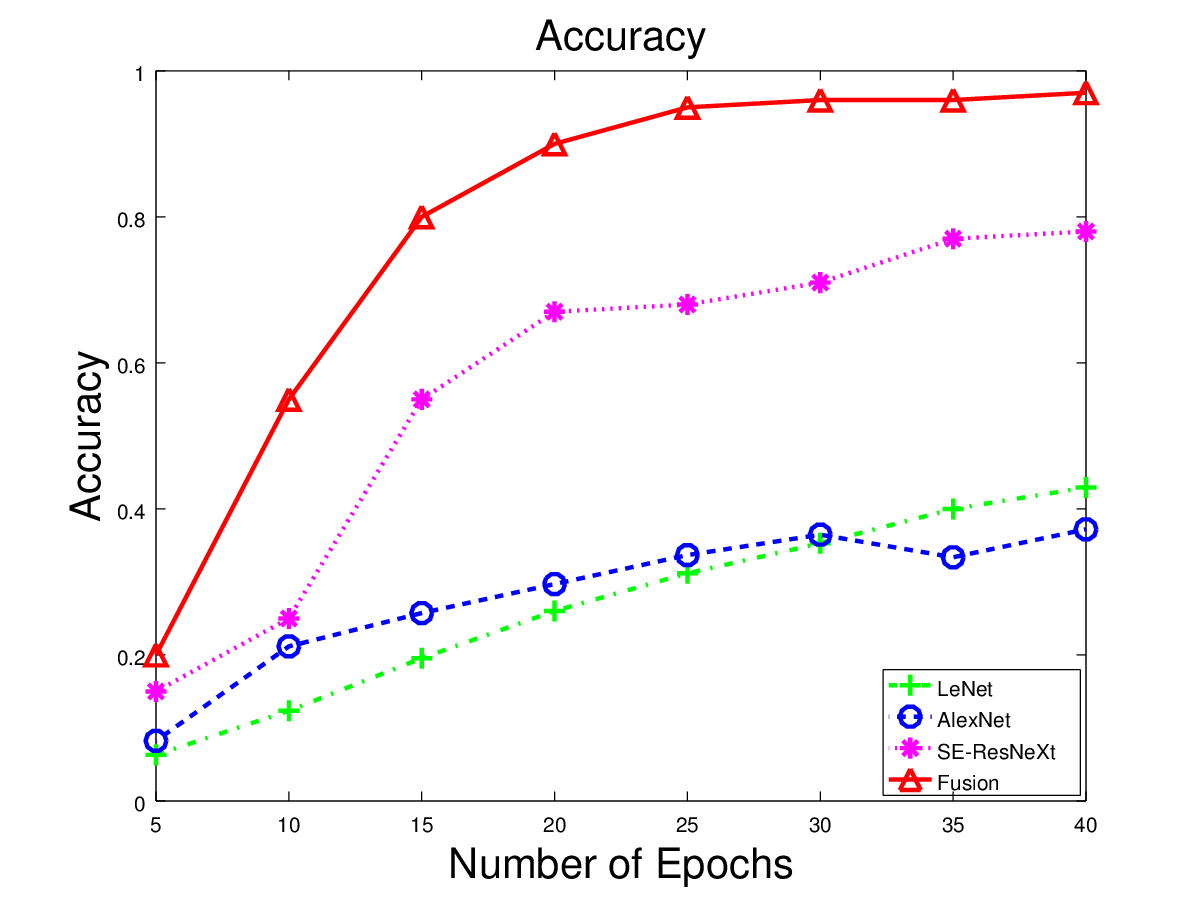}
    \caption{Accuracy}
    \label{Accuracy}
  \end{minipage}
  \hfill
  \begin{minipage}[b]{0.49\textwidth}
    \includegraphics[width=\textwidth, height=6cm]{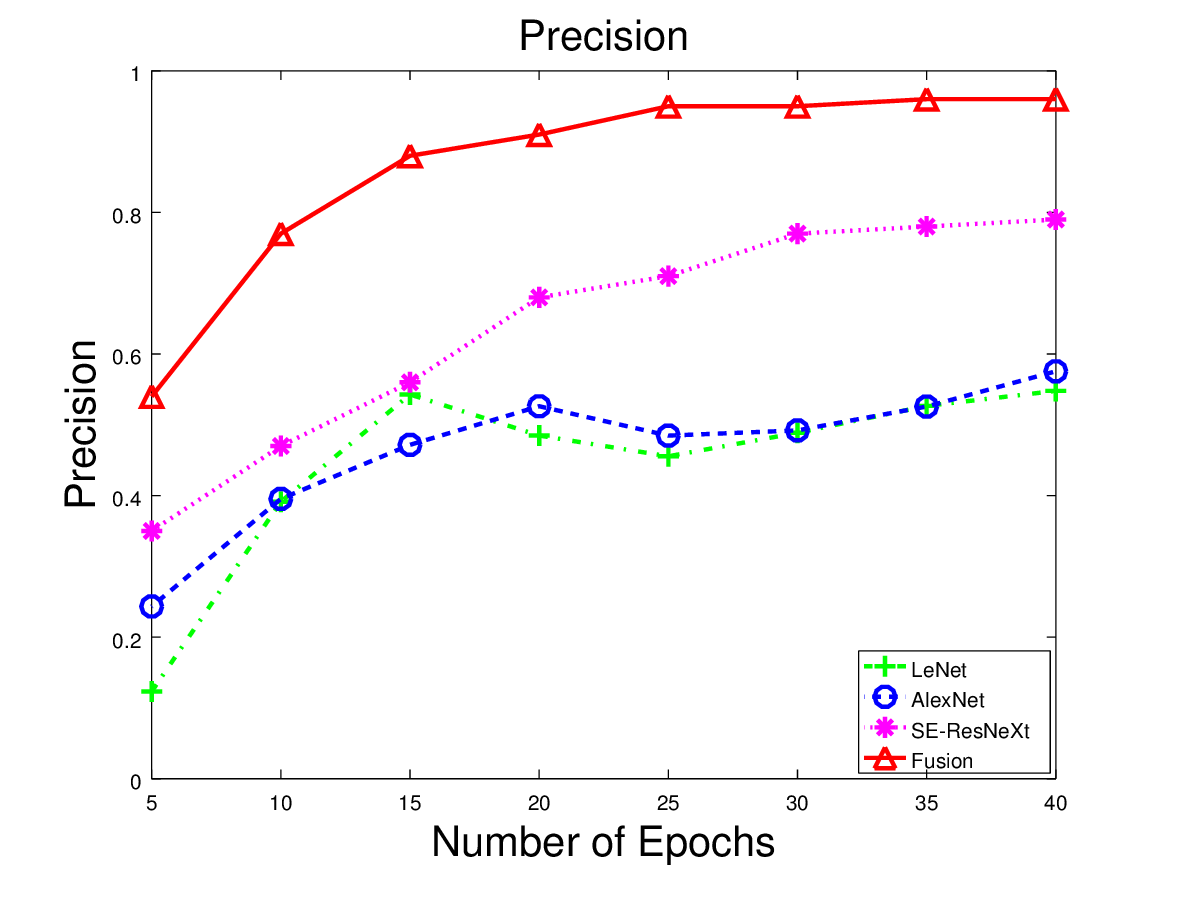}
    \caption{Precision}
    \label{Precision}
  \end{minipage}
    \begin{minipage}[b]{0.49\textwidth}
    \includegraphics[width=\textwidth, height=6cm]{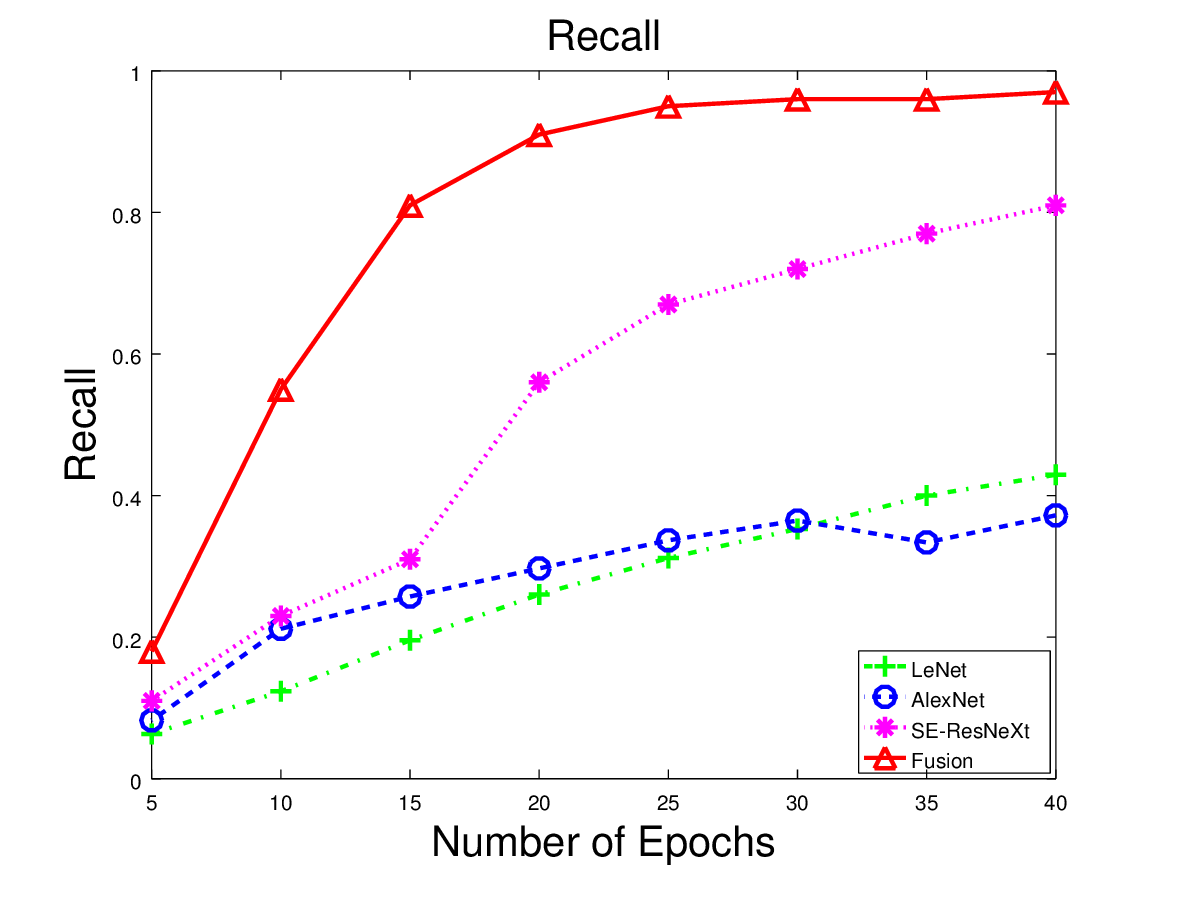}
    \caption{Recall}
    \label{Recall}
  \end{minipage}
    \begin{minipage}[b]{0.49\textwidth}
    \includegraphics[width=\textwidth, height=6cm]{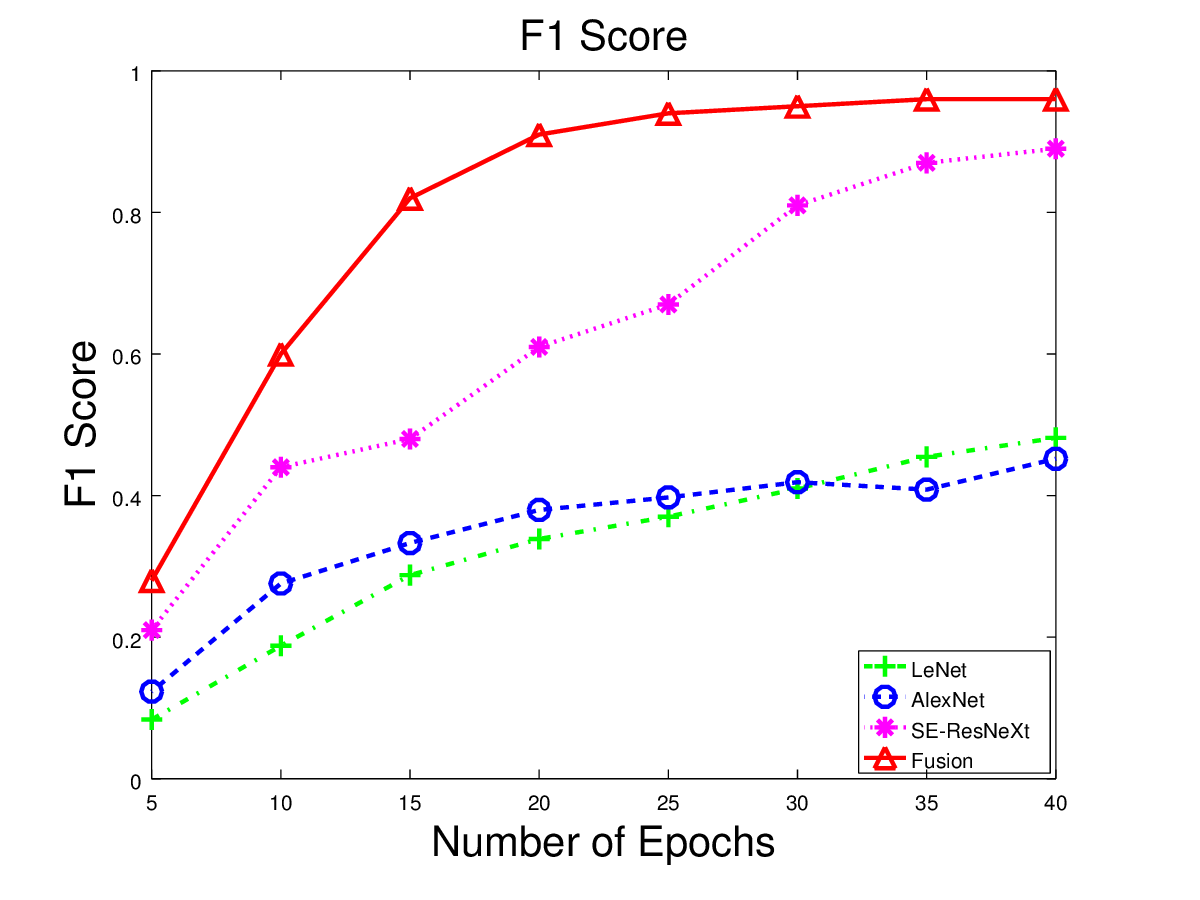}
    \caption{F1 Score}
    \label{F1 Score}
  \end{minipage}
\end{figure*}

We have also compared the proposed system with  well-known deep learning models, i.e. AlexNet, LeNet and SE-ResNeXt. Figure \ref{Accuracy}, Figure \ref{Precision}, Figure \ref{Recall} and Figure \ref{F1 Score} depict and compare performance improvements of the proposed system with the AlexNet, LeNet and SE-ResNeXt models. As it can be seen from Fig.~\ref{Accuracy},  the proposed orientation fusion approach provides much higher accuracy rates, as compared to AlexNet, LeNet and SE-ResNeXt. From the very start, at iteration number 5, the accuracy of the system is recorded to be 0.2, while the accuracy of AlexNet, LeNet and SE-ResNeXt was below 0.2. The accuracy kept on improving with the increase in epochs. At epoch number 25, a significant improvement can be observed in the graph, as compared to the other three models, which kept on improving till the last epoch.

A similar kind of behavior can be observed in the precision, recall and F1 scores, as shown in Figures \ref{Precision}, \ref{Recall} and \ref{F1 Score}. The precision of the system started from 0.55 (at epoch 5) and showed a linear improvement over increasing number of epochs. Rapid improvements have been observed till epoch number 25 in the precision curve, which kept on improving gradually till the last epoch. The recall and F1 score curves depict a similar trend in their curves, and show significant improvements as compared to AlexNet, LeNet and SE-ResNeXt. The AlexNet performed a bit better than the LeNet till epoch number 30, but it could not get better than score 0.3. A drop in the recall curve for AlexNet has been observed after 30 epochs. The same trend was observed for the F1 score curve. 

Given the results, we can conclude that the proposed orientation-fusion approach exhibits significant improvements over the three benchmark models, on a challenging dataset. The images present in the publicly available Yale face database have significant variations in expressions, pose and illumination conditions. The illumination conditions impose effects from different angles. Furthermore, expressions of the individuals present in the database varied from normal to happy, sad and sleepy as well. It was observed from the results that the system reported in this paper is superior than existing systems in tackling these challenges. 

We believe that the reason behind these improvements is that the illumination effects are present in low frequency components of the spectrum. The EMD separates the images into individual intrinsic mode functions in the decreasing order. It then becomes easier to discard the low frequency components from the image, and retain only the high frequency components. The fusion of the two intrinsic mode functions containing the highest frequencies is sufficient  to correctly classify most of the training samples, with high accuracy rate and precision. \\

\textbf {\textit {Scalability of the System.}} The cloud infrastructure used to execute the visual object recognition system includes one master node working with eight worker nodes. The cloud infrastructure helps to parallelize the proposed object-recognition approach by executing the chunks of data (subsets) on multiple worker nodes. The total dataset size ranges from ten to a hundred gigabytes. The dataset is then divided into subsets, and each subset of data is further divided into a number of mini-batches. Each worker processes each mini-batch. The dataset is exported to the distributed file system in a batched and serialized form.



During the training process of visual object recognition, initial configurations and parameters are loaded into memory. The master node then distributes the subsets of data to worker nodes, along with the initialization parameters. Each model on the worker node is trained on different shards of the data in the form of mini-batches. The results are then averaged using parameter averaging on the master node. This approach is quite useful in our case as the number of worker nodes is small and the parameters that are to be estimated are also small. The parameter averaging is performed by obtaining the gradient of each mini-batch from all the nodes. After the completion of the training process, the master node holds the fully trained model.



We have mainly focused on three measures in terms of scalability: i) the total time needed to transfer different sizes of the datasets to the cloud storage; ii) the average execution time with varying dataset sizes; and iii) the average time with different nodes. 

\begin{figure*}[!tbp]
  \centering
  \begin{minipage}[b]{0.47\textwidth}
    \includegraphics[width=\textwidth, height=6cm]{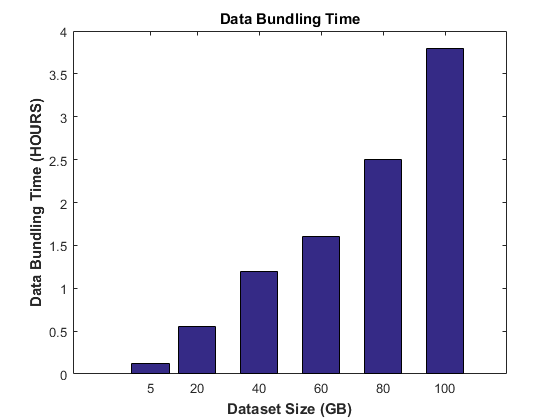}
    \caption{Data Bundle Time}
    \label{Data Bundle Time}
  \end{minipage}
  \hfill
  \begin{minipage}[b]{0.49\textwidth}
    \includegraphics[width=\textwidth, height=6cm]{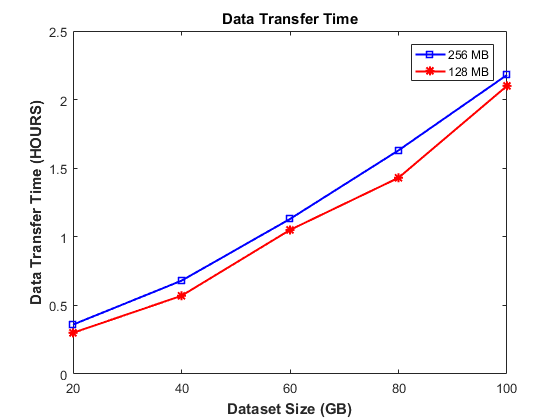}
    \caption{Data Transfer Time}
    \label{Data Transfer Time}
  \end{minipage}
    \begin{minipage}[b]{0.49\textwidth}
    \includegraphics[width=\textwidth, height=6cm]{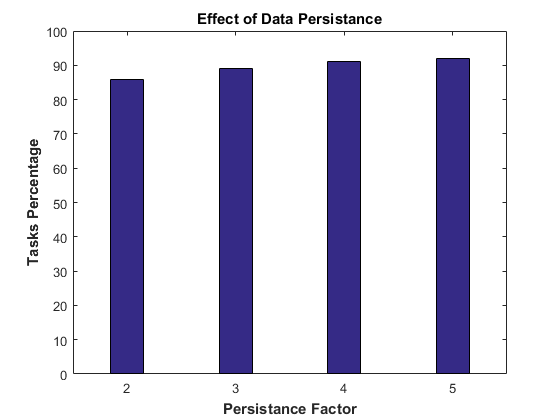}
    \caption{Effect of Data Persistence}
    \label{Persistence}
  \end{minipage}
    \begin{minipage}[b]{0.49\textwidth}
    \includegraphics[width=\textwidth, height=6cm]{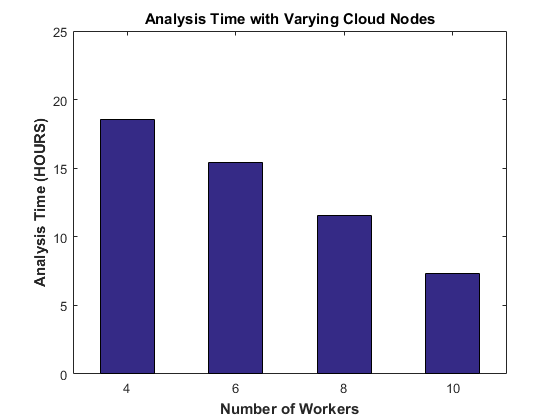}
    \caption{Analysis Time}
    \label{Analysis Time}
  \end{minipage}
\end{figure*}

The total size of the dataset is varied from 5GB to 100GB to measure the scalability of the system. This dataset consists of the image database as well as a number of video streams consisting of multiple subjects. The number of decoded frames are dependent on the size of each video stream. We have further used a batch process in order to bundle the large amount of individual decoded frames and images. The iterative map-reduce framework works better with the large bundled data as compared to individual small chunks of data. The time required by the batch process to bundle the data depends on the size of the total dataset. Fig.~\ref{Data Bundle Time} shows the time taken by the batch process, on different sizes of the dataset. For a dataset size ranging from ten to a hundred gigabytes, it took from 0.26 to 3.9 hours.     

The data transfer time from local storage to cloud depends on the amount of data that is being transferred. There are two main factors that contribute to the total data transfer time, i.e. bandwidth and block size. We have measured the execution time with different data sizes. The data transfer time for each dataset size is shown in Fig.~\ref{Data Transfer Time}. The data transfer time is directly proportional to the size of data that is being transferred. It requires almost 0.36 hours to transfer 20 GB data. It takes almost two hours and eighteen minutes to transfer 100 GB to the cloud. An increase in the amount of dataset size will also increase the data transfer time but this is a one-time process as the data could be retained in the cloud storage for later use.  


We have varied nodes of the cloud infrastructure and measured the average execution time. Multiple experiments have been performed with an increasing number of nodes. This gives an estimate of the amount of time required for system execution with multiple nodes. The analysis time incurred by the system with multiple nodes is plotted in Fig.~\ref{Analysis Time}. This helps to have an idea as to how many nodes should be required to execute the system, within a reasonable time frame. A decreasing trend in the analysis time is observed by increasing the number of workers on each node.





\section{Conclusion and Future work}

In this paper, an illumination- and expression--invariant video analytics system for visual object recognition has been proposed, to tackle challenging datasets. We introduce a novel feature-fusion strategy based on the orientation component of intrinsic mode functions (IMF). The IMFs are generated by leveraging bi-dimensional empirical mode decomposition. Also, the  first-order Reisz transform is exploited to produce an orientation component. The fused IMFs are further analyzed using CNNs.

The effectiveness of the proposed system is demonstrated by experimentation on publicly available datasets and compared with two existing benchmark models namely AlexNet and LeNet. It is observed that the orientation component of the objects leads to a comparatively higher accuracy of 93\%. Our proposed feature fusion strategy of the orientation spectra further improves the accuracy to 97\%. The precision and recall are found to be 98\%
and 97\%, respectively. In summary, our proposed system proved to be highly accurate and precise and outperformed AlexNet, LeNet and SE-ResNeXt under uncontrolled conditions.

The proposed system can be beneficial for a number of machine vision tasks such as object detection, object classification and object tracking. It can be used in a number of applications such as video surveillance, person monitoring, traffic monitoring and tracking, to mention but a few. These applications have their importance in smart cities, safe cities and IoT. It can also be useful in other domains such as medical image processing and satellite imagery.

In terms of limitations, the computation cost of the IMFs generated through the EMD process is relatively high, which makes the overall system resource- and compute--intensive. The current system also lacks a feed-back process for online training with new incoming samples of the training dataset. Further, more complex challenges such as rotation and translation variance are yet to be tackled.

In the future, we aim to enhance and optimise the capabilities of the proposed system so as to further improve on  scalability challenges as well as the rotation and translation variance. Novel approaches for online information extraction from dual and multi-data streams will be exploited. We also aim to execute the proposed system on multiple nodes of a GPU-based cloud
infrastructure. The latter will help manage the complexity of the proposed model and facilitate comparative experiments on even bigger benchmark datasets.

We aim to propose a high performance video stream processing platform using the proposed orientation fusion based approach for visual object tracking. The high--performance platform equipped with GPUs will aid to rapidly compute and fuse orientation components of large--scale video datasets. This platform will help to track multiple objects from multiple video streams containing real--life challenges such as illumination variance and blur effects.

\label{}









\bibliography{bibtex/bib/sample.bib}{}
\bibliographystyle{unsrtnat}

\end{document}